\documentclass[letterpaper, 10 pt, conference]{ieeeconf}
\IEEEoverridecommandlockouts
\usepackage{amssymb}
\usepackage{amsmath}
\usepackage[draft]{minted}
\usepackage[noadjust]{cite}
\usepackage{listings}
\usepackage{graphicx}
\usepackage{macros}
\usepackage{url}
\usepackage{svg}
\usepackage{booktabs}
\usepackage[colorinlistoftodos]{todonotes}
\usepackage{hyperref}
\usepackage{ifthen}
\definecolor{codegreen}{rgb}{0,0.6,0}
\definecolor{codegray}{rgb}{0.5,0.5,0.5}
\definecolor{codepurple}{rgb}{0.58,0,0.82}
\definecolor{backcolour}{rgb}{0.95,0.95,0.92}
 
\lstdefinestyle{mystyle}{
    language=Python,
    backgroundcolor=\color{backcolour},   
    commentstyle=\color{codegreen},
    keywordstyle=\color{magenta},
    numberstyle=\tiny\color{codegray},
    stringstyle=\color{codepurple},
    basicstyle=\ttfamily\footnotesize,
    breakatwhitespace=false,
    breaklines=true,
    captionpos=b,
    keepspaces=true,
    numbers=left,
    numbersep=5pt,
    showspaces=false,
    showstringspaces=false,
    showtabs=false,
    tabsize=2,
    basicstyle=\fontsize{6}{7}\selectfont\ttfamily,
}
 
\newboolean{commentsOn}
\setboolean{commentsOn}{False}

\ifthenelse{\boolean{commentsOn}}{\newcommand{\bardh}[1]{{\color{blue}{\bf BH:} #1}}}
{\newcommand{\bardh}[1]{}}

\DeclareMathOperator*{\argmin}{arg\,min}
\DeclareMathOperator{\R}{\mathbb R}
\newcommand{\bb}{\boldsymbol}
\newtheorem{Definition}{{\bf Definition}}

\title{\LARGE \bf 
\textsc{CBFkit}: A Control Barrier Function Toolbox for Robotics Applications
}

\author{Mitchell Black, Georgios Fainekos, Bardh Hoxha, Hideki Okamoto, Danil Prokhorov% <-this % stops a space
\thanks{The authors are listed in alphabetical order and are with Toyota Motor North America R\&D, 
        Ann Arbor, MI 48105, USA. email:
        {\tt\small first.last@toyota.com}}%
}

\begin{document}

\maketitle
\thispagestyle{empty}
\pagestyle{empty}

\begin{abstract}
This paper introduces \cbfkit{}, a Python/ROS toolbox for safe robotics planning and control under uncertainty. The toolbox provides a general framework for designing control barrier functions for mobility systems within both deterministic and stochastic environments. It can be connected to the ROS open-source robotics middleware, allowing for the setup of multi-robot applications, encoding of environments and maps, and integrations with predictive motion planning algorithms. Additionally, it offers multiple CBF variations and algorithms for robot control. The CBFKit is demonstrated on the Toyota Human Support Robot (HSR) in both simulation and in physical experiments. 
\end{abstract}

% \listoftodos

% \noindent \textbf{Toolbox Features}
% \todo[inline, color=blue!40]{\checkmark\textbf{MB}: Generate new sim tutorial}
% \todo[inline, color=blue!40]{\checkmark\textbf{MB}: Single and Monte Carlo examples}
% \todo[inline, color=blue!40]{\textbf{MB}: Additional forms of CBF (adaptive, etc.)}

% \noindent \textbf{Unittests}
% \todo[inline, color=green!100]{\checkmark Finish controller suite}
% \todo[inline, color=green!100]{\checkmark Finish optimization suite}
% \todo[inline, color=green!100]{Lower level function tests (math funcs, etc.)}

% \noindent \textbf{Tutorials / Benchmarks}
% \todo[inline, color=purple!50]{\checkmark\textbf{MB}: Single trial reach-avoid}
% \todo[inline, color=purple!50]{\textbf{MB}: Monte Carlo reach-avoid}
% \todo[inline, color=purple!50]{\checkmark \textbf{MB}: Using cbfkit in a new repo (tutorial)}

% \noindent \textbf{Documentation}
% \todo[inline, color=yellow!80]{\textbf{BH}: Detailed README}
% \todo[inline, color=yellow!80]{\textbf{BH}: Docstring for every file}
% \todo[inline, color=yellow!80]{\textbf{MB}: Docstring for every function}

% \noindent \textbf{Integration}
% \todo[inline, color=red!80]{\textbf{MB / BH}: Migrate internal repo to public}
% \todo[inline, color=red!80]{\textbf{BH}: Verify passes mypy/integration tests}
% \todo[inline, color=red!80]{\checkmark\textbf{BH}: Confirm Docker functionality}

\section{Introduction}

Automated mobility systems are typically deployed in safety-critical applications, or those for which failure may result in loss of life, damage to the system, or damage to its environment.
Such applications require higher levels of assurance to establish safety and reduce risks.
Safety standards, such as ISO 22737 \cite{iso22737_2021},  ISO 34502 \cite{iso34502_2022}, ISO/TS 15066 \cite{ISO15066_2016}, etc, have been developed to capture requirements, operational domains, and testing procedures for mobility systems of various levels of automation.  
Simulation based testing from requirements \cite{TuncaliEtAl2020tiv} can be utilized to assess system performance and its relation to the real system performance \cite{FremontEtAl2020itsc}; 
however, by definition, testing-driven approaches cannot address the completeness question:
Did we create enough test cases to validate the performance of the system in the whole operating domain?

To make progress toward addressing this question, at the very least, we need to develop control methods which can guarantee safety.
Control Barrier Functions (CBF) \cite{AmesEtAl2019ecc} provide such a mechanism for a wide variety of applications from (semi-)automated driving \cite{AmesGT2014cdc,XiaoEtAl2021iccps,HeZZS2021acc,MolnarEtAl2022ifac} to arm manipulators \cite{ShawCortezVD21icra,SingletaryEtAl2022ral,FerragutiEtAl2020ras} to multi-agent coordination \cite{ParwanaMP2022cdc,LindemannD2019csl,MehmoodEtl2023jas,EmamEtAl2022tr}. CBFs enforce a barrier-like inequality condition on the control input to the system and, in doing so, may be used to render a set of states satisfying a given constraint forward invariant.

CBF-based control is characterized by some distinct advantages that make it a good candidate for control within a safety-first development framework.
First, CBFs provide a mechanism to guarantee safety similarly to how Lyapunov theory guarantees stability.
In many cases, CBFs can also be combined with control Lyapunov functions (CLFs) to both stabilize the system and guarantee safety, either as pairs of affine constraints in an optimization-based controller \cite{AmesGT2014cdc,GargP2019cdc} or via jointly synthesized Lyapunov-Barrier functions
\cite{BraunKZ2019ifac_ncs,WuEtAl2019automatica}.
The definition of the safe operating region under the influence of a CBF (forward invariant) is flexible and it provides an unambiguous way to document the operational domain for safety certification.
In addition, many CBF-based control methods can be very efficiently implemented online for control-affine system models and some classes of hybrid systems \cite{MengL2023nahs},
% CBF methods apply to systems that can be modeled using control affine (stochastic) systems,
which are typically general enough to capture many true systems of interest.
Finally, CBFs provide a versatile theoretical framework for control design using analytical formulations \cite{BraunKZ2019ifac_ncs,WielandA2007ifac_ncs} or optimization based control \cite{AmesGT2014cdc,YaghoubiFS2020itsc}, and have even shown promise in randomized methods \cite{AhmadBT2022cdc,MajdEtAl2021iros} and in predictive control both as a stand-alone framework \cite{BlackJSP2023acc,BreedenP2022cdc} and as constraints in Model Predictive Control (MPC) \cite{ZengZS2021acc,LafmejaniBF2022iros,GrandiaEtAl2020rss}.

In this work, we build upon the momentum of CBF research to develop an open source publicly%
\footnote{\url{https://github.com/bardhh/cbfkit.git}}
available Python package \cbfkit{} for CBF-based control, which can be deployed within the Robot Operating System (ROS)%
\footnote{\url{http://www.ros.org/}}.
Our objective is to build, maintain, and support a modular extendable toolbox that encompasses some of our recent research in risk-aware CBF control \cite{YaghoubiEtAl2021csl,YaghoubiEtAl2021cdc,MajdEtAl2021iros,BlackFHPP2023icra} and beyond \cite{YaghoubiFS2020itsc,LafmejaniBF2022iros,AgrawalP2022lcss,BreedenGP2021lcss,BlackP2022ifacwc}. 
We take a functional programming approach where the user needs to instantiate functions representing a model/system and a controller.
The controller itself can be model-based, where the model of the system is used for control, or model free.
In the latter case, \cbfkit{} could be interfaced with falsification/design optimization Python tools such as $\Psi$-\textsc{TaLiRo} \cite{ThibeaultEtAl2021fmics} and TLTk \cite{cralley2020tltk} to identify control parameters that optimize mission level requirements.
In addition, our control framework can be used in conjunction with requirements online monitoring tools such as RTAMT \cite{NickovicY2020atva}.
Finally, we demonstrate the \cbfkit{} framework working with the Toyota Human Support Robot (HSR) \cite{YamamotoEtAl2019hsr} in both real and simulated experiments.

{\bf Contribution:}
We present the first publicly available Python toolbox, called \cbfkit{}, for CBF-based control in ROS.
\cbfkit{}  is modular and extendable with the ultimate goal of being the go-to-software library for all CBF needs. 
The toolbox not only contains control examples for Toyota HSR, but also tutorials for newcomers to the CBF-based control methods. 
% In addition, the toolbox includes multiple implementations of CBF and CLF controllers from literature \todo{@mitchell, can you please add some citations}.

\section{Supported Models and Control Design Problems}

Our goal for \cbfkit{} is to be an all encompassing tool for CBF-based feedback control design.
As such, \cbfkit{} supports a number of different classes of control-affine models (model of system $\Sigma$):

\begin{enumerate}
\item Deterministic, continuous-time Ordinary Differential Equations (ODE): 
\begin{equation}
\label{sys:ode}
    \dot{x}=f(x)+g(x)u,
\end{equation}
where $x\in \mathcal X\subset \mathbb{R}^{n}$ is the system state,  $u\in \mathcal U \subset \mathbb{R}^{m}$ is the control input, $f:\mathbb{R}^{n} \rightarrow \mathbb{R}^{n}$ and $g:\mathbb{R}^{n} \rightarrow \mathbb{R}^{n \times m}$ are locally Lipschitz functions, and $x(0) \in \mathcal X_0 \subseteq \mathcal X$ is the initial state of the system.

\item Continuous-time ODE under bounded disturbances:
\begin{equation}
\label{sys:ode_uncertain}
\dot{x}=f(x)+g(x)u+Mw,
 \end{equation}
where $w\in \mathcal W$ is the disturbance input,  $\mathcal W$ is a hypercube in $\mathbb{R}^{l}$, and $M$ is a $n\times l$ zero-one matrix with at most one non-zero element in each row.

\item Stochastic differential equations (SDE):
\begin{equation}
\label{sys:sde}
 \mathrm{d}{x} = \big(f({x}) + g({x}){u}\big)\mathrm{d}t + \sigma({x})\mathrm{d}{w}
 \end{equation}
where $\sigma: \R^n \rightarrow \R^{n \times q}$ is locally Lipschitz, and bounded on $\mathcal X$, and ${w} \in \R^q$ is a standard $q$-dimensional Wiener process (i.e., Brownian motion) defined over the complete probability space $(\Omega, \mathcal{F}, P)$ for sample space $\Omega$, $\sigma$-algebra $\mathcal{F}$ over $\Omega$, and probability measure $P: \mathcal{F} \rightarrow [0,1]$. 

\end{enumerate}

The above models are further assumed to be memory-less in that the instantaneous dynamics are independent of the time history of the system. Discrete-time variants can also be considered for, e.g., reinforcement learning methods or MPC.
For example, in the literature \cite{ZengZS2021acc,LafmejaniBF2022iros,GrandiaEtAl2020rss}, it is assumed that the ODE model in discrete time takes the  form:
\[ x_{i+1} = x_i + \int_{t_i}^{t_{i+1}} \big[f(x(t))+g(x(t))u_i\big] dt \]
where the integral can be approximated either analytically or with numerical methods.

In certain practical applications, not all the states of the system may be observable. 
In such scenarios, we may assume that a state vector $y$ is observable.
For example, in the case of SDE, we may assume:
\[ dy = Cx \mathrm{d}t + D \mathrm{d}v \]
where $C \in \mathbb R^{p \times n}$, $D \in \mathbb R^{p \times r}$, and $v \in \R^r$ is a standard Wiener process.

The framework of CBF-based control can provide safety guarantees by enforcing a forward invariant set.
Namely, if the system starts in a safe state, then it should always stay in the safe set.

\begin{Definition}
{[Set Invariance \cite{blanchini1999set}]}
\label{def2}
A set $S \subseteq \mathbb{R}^n$ is forward invariant w.r.t the system $\Sigma$ iff for every $x(0) \in S$, its solution satisfies $x(t)\in S$ for all $t \geq 0$. 
\end{Definition}

From a usability perspective, it may be easier to specify the unsafe set of states $\bar S$. 
The unsafe set of states $\bar S$ for a model $\Sigma$ can be defined using a locally Lipschitz function 
$h:  \mathbb{R}^n\rightarrow \mathbb{R}_+$ as 
\begin{equation}\label{eq:unsafe} % \nonumber
% $
\bar S =: \{ x \;|\;h(x) < 0 \}.  
% $
\end{equation}
It follows that the safe set of states is 
\begin{equation}\label{eq:safe} % \nonumber
%$
S = \mathbb{R}^n \backslash \bar S = \{ x \;|\;h(x) \geq 0 \}.
%$
\end{equation}
In \cbfkit{}, the end-user is free within the framework provided to specify  such a function (or functions) $h$, the CBF condition for which is then automatically evaluated for the provided model given the known states and inputs.

% \todo[inline]{@B or @M: we need 1-2 sentences on how we define h above in the toolbox}

Given an unsafe set $\bar S$ and depending on the type of the model $\Sigma$ under consideration, we have two different types problems that we consider:
\begin{enumerate}
    \item In case of system models (\ref{sys:ode})- (\ref{sys:ode_uncertain}), the goal is to design a control $u(t)$ such that the state of the system $x(t)$ never enters the unsafe set $\bar S$ for any time $t\geq 0$.

    \item In case of system models (\ref{sys:sde}), the goal is to design a control feedback $u(t)$ such that the probability that the state of the system $x(t)$ enters the unsafe set $\bar S$ is upper bounded by a probability $\bar p$ over a bounded time interval $[t,t+T]$ for some finite time $T$.
\end{enumerate}

In \cbfkit{}, we provide solutions (feedback controllers) to the above two problems using Quadratic Program formulations as in, e.g., \cite{BlackJSP2023acc,AmesGT2014cdc} for model (\ref{sys:ode}), \cite{YaghoubiFS2020itsc,jankovic2018robust} for model (\ref{sys:ode_uncertain}), and \cite{YaghoubiEtAl2021csl,YaghoubiEtAl2021cdc,BlackFHPP2023icra} for model (\ref{sys:sde}).

% \todo[inline]{GF: We should only mention SDE and MPC if there in the horizon for implementation before potential publication
% MB: Definitely SDE -- I would say probably MPC before publication but no guarantees on completing it before the paper is reviewed.}

\section{CBFkit Toolbox}
 
The Toolbox is developed in Python and is designed to connect to Robot Operating System (ROS). 

\subsection{Software Design}

The toolbox is built on functional programming principles, emphasizing data immutability and programmatic determinism. 
% Within this framework, we provide functions responsible for carrying out tasks pertinent to the following modules: \texttt{codegen}, \texttt{controllers}, \texttt{estimators}, \texttt{integration}, \texttt{modeling}, \texttt{optimization}, \texttt{ros}, \texttt{ros2}, \texttt{sensors}, \texttt{simulation}, and \texttt{sensors}.
This programming paradigm ensures that system states are not altered in-place; instead, functions return new states, thereby enhancing code reliability, maintainability, and facilitating debugging and testing.

The toolbox architecture is centered around a collection of pure functions that represent the core operations required for simulation and control of dynamical systems. These functions are organized into modules corresponding specific functionalities \texttt{dynamics}, \texttt{controller}, \texttt{estimator}, \texttt{integrator}, \texttt{perturbation}, and \texttt{sensor}. Each function is designed to take input parameters and return new outputs without side effects, adhering to the principles of functional programming.

The architecture replaces traditional object-oriented constructs, such as classes and inheritance, with function compositions and higher-order functions. For instance, system dynamics and control strategies are defined through composable functions rather than through methods of a \texttt{System} class. This approach enables more flexible and reusable code, as functions can be easily combined and reused across different parts of the system without the tight coupling introduced by class inheritance.

For improved usability, we provide template functions for \texttt{dynamics} (i.e., given a state $x$ compute the control-affine dynamics $f(x)$ and $g(x)$), \texttt{controller} (i.e., given a time $t$ and state $x$ compute the control input $u(t, x)$).

Function currying is employed to refine functionality; for instance, a \texttt{cbfcontroller} function, a specialized form of \texttt{controller}, initially requires multiple control parameters during setup. Post-initialization, the \texttt{controller} function consistently requires time $t$ and state vector $x$ as arguments. Similarly, while a \texttt{plant} may be initialized with numerous parameters, its \texttt{dynamics} function invariably accepts time $t$ and state vector $x$ as inputs.

The toolbox is managed using Poetry\footnote{\url{https://python-poetry.org}} and features both a Docker container and a VsCode devContainer. This setup expedites development and tutorial execution, ensuring a streamlined process and consistent user environment.

\subsection{Arguments/Return Types for Important Template Functions}

The functional design is reflected in the detailed specification of input and output types for key functions within the toolbox, as outlined in the Table \ref{tb:functions}. This explicit type declaration aids in understanding the flow of data through the system and ensures that functions can be composed safely and predictably.

% \todo{\textbf{BH}: Re-do (functional programming)}

% The \texttt{System} class enables the definition of system dynamics and also provides a method for simulating the system. 

% This class is extended by the \texttt{Simulated} class for systems that can be simulated, and the \texttt{Hardware} class for systems simulated through hardware or physical robot applications. 
% Additionally, the \texttt{Simulated} class enables the definition of stochastic systems by defining noise parameters, noise distribution, and other statistical terms. The extension of the \texttt{System} class to the \texttt{Simulated} and \texttt{Hardware} classes is useful for practical applications since the ground truth of the system is not always available and the system states may not be observable.

% \begin{figure*}[t]
% \centering
% \resizebox{13cm}{!}
% \includesvg{fig/functions.svg}
% \caption{Caption.}
% \label{fig:functions}
% \end{figure*}

\begin{figure}[t]
  \centering
  \includegraphics[width=\columnwidth]{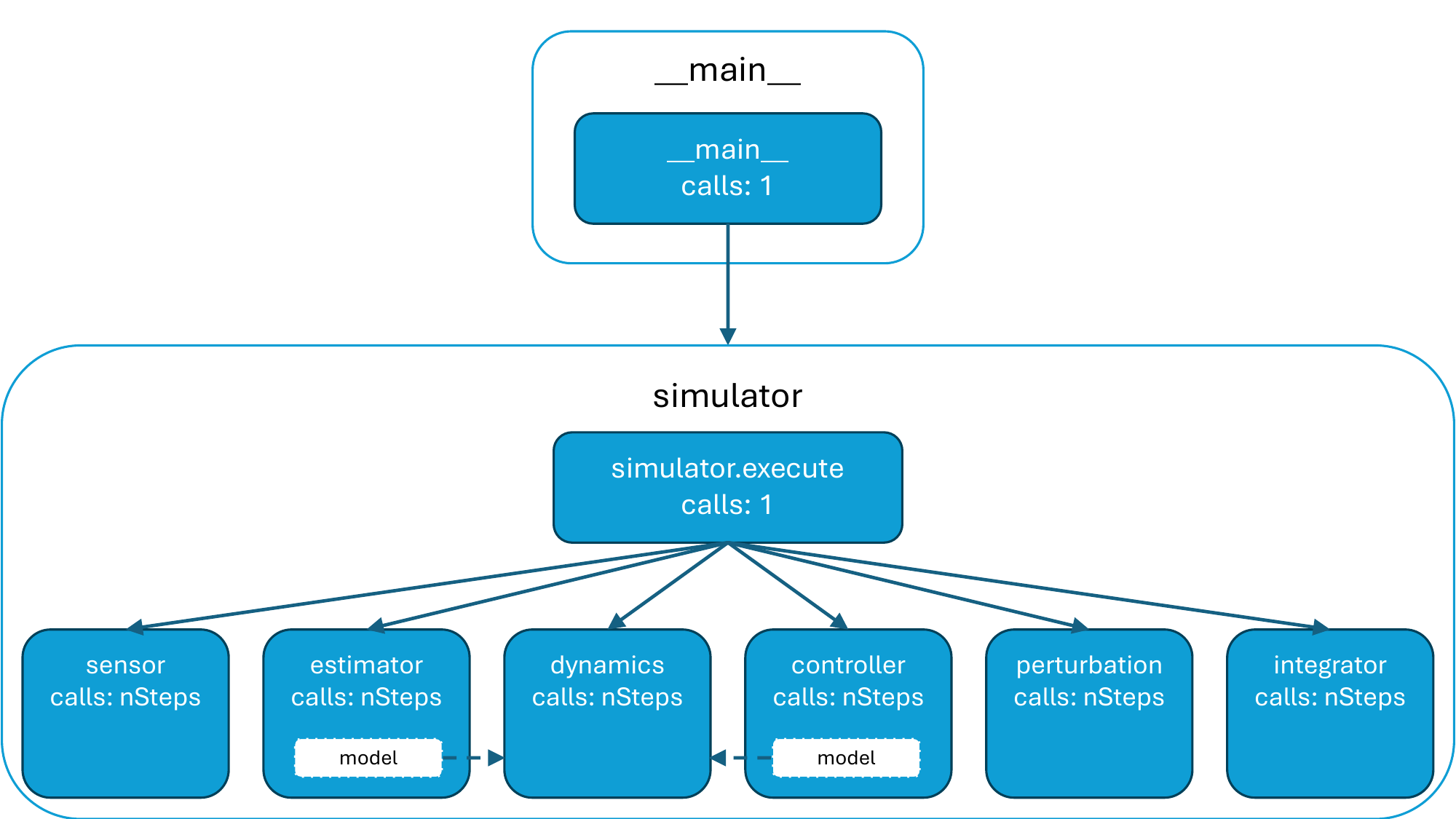}
  \caption{Call graph depicting relations between function calls important for building an example simulation, i.e., \texttt{sensor}, \texttt{estimator}, \texttt{dynamics}, \texttt{controller}, \texttt{perturbation}, and \texttt{integrator},  using \cbfkit{}.}
  \label{fig:function_graph}
  \vspace{-5px}
\end{figure}

% \todo[inline]{\textbf{MB}: Add a table to describe the input/output types of the dynamics, perturbation, integrator, controller, sensor, and estimator functions.}

\begin{table*}[ht]
\centering
\caption{Arguments/Return Types for Important Template Functions}
\label{tb:functions}
\begin{tabular}{l|lc|lc}
\toprule
\multicolumn{1}{c}{\textbf{Function}} & \multicolumn{1}{c}{\textbf{Arguments}} & \multicolumn{1}{c}{\textbf{Types}} & \multicolumn{1}{c}{\textbf{Returns}} & \multicolumn{1}{c}{\textbf{Types}} \\
\midrule
controller & $t$, $x$ & float, Array[$n$] %(float,array)  \rightarrow array, dict$ 
& $u$, data & Array[$m$], Dict[str, Any] \\
dynamics & $t$, $x$ & float, Array[$n$] & $f$, $g$ & Array[$n$], Array[$n$, $m$] \\
estimator & $t$, $y$, $z$, $u$, $c$ & float, Array[$p$], Array[$n$], Array[$m$], Array[$p$, $p$] & $z$, $c$ & Array[$n$], Array[$p$, $p$] \\
integrator & x, $\dot x$, $\mathrm{d}t$ & Array[$n$], Array[$n$], float & $x$ & Array[$n$] \\
perturbation & $x$, $u$, $f$, $g$ & Array[$n$], Array[$m$], Array[$n$], Array[$n$, $m$] & pert & Callable[[jax.random.PRNGKey], Array[$n$]] \\
sensor & $t$, $x$, $\sigma$, key & float, Array[$n$], Array[$n$, $n$], jax.random.PRNGKey & $y$ & Array[$p$] \\
\bottomrule
\end{tabular}
\vspace{2mm}

Note: arguments and return variable names are taken from the code, and Array[$k$] denotes a \texttt{jax.numpy.array} with data type \texttt{float} of length $k$.
\end{table*}

% The \texttt{Controller} class is used to define the feedback controller. 
% The \texttt{Controller} class is extended by \texttt{Model-based} and \texttt{Model-free} classes. The \texttt{Model-free} controller class is extended by the \texttt{PID} and \texttt{Reinforcement Learning (RL)} classes. 
% The \texttt{Model-based} class is extended by the \texttt{MPC}, \texttt{CBF}, and \texttt{Lyapunov} classes which, as the name implies, require a model as well. This enables the user to define a wide range of controllers for their system. The interaction of a system class with a CBF controller is shown in in Figure.~\ref{fig:object_interaction}.

% The \texttt{System} class includes a \texttt{step()} method which enables the user to simulate the system dynamics. The \texttt{step()} method takes the system state, controller output, and time step as inputs and returns the next system state. For ROS applications, the closed-loop interaction between the system and controller is handled by a ROS adapter. A ROS node subscribes to the controller output and publishes the system state.

% \todo[inline]{Specify which models.}
The toolbox provides users with a library of example models, such as double integrator, unicycle, bicycle models and omnidirectional robots. It also includes example scripts that show how the toolbox can be used for various models and controllers. Also, some handy utility functions are included which can help store, load, and visualize system behaviors.

\subsection{CBF Implementation}

The implementation of a CBF-based controller utilizes the ego system dynamics, additional agents' dynamics, and the environment to synthesize a controller to enforce that a region of the state space satisfying a set of safety constraints is rendered forward invariant. There are two main steps in the control synthesis process: 1) safety constraint generation and 2) controller design. 

% The safety constraints guarantee that the system will stay within a safe set of states. 
% Typically, these are derived from a barrier function as in Eq. (\ref{eq:safe}). 
%, which is zero on the safe set and its derivative is positive outside the safe set. 
In order to generate the safety constraints, a barrier function as in \eqref{eq:safe} is related to the system inputs via mathematical derivation. A key feature of the toolkit is the use of the auto-differentiation capabilities of JAX~\cite{jax2018github} for computation of the derivative of the barrier function. For complex dynamics, this derivation can be computationally challenging using symbolic toolboxes such as SymPy~\cite{10.7717/peerj-cs.103}. 
% Instead, we utilize JAX for auto-differentiation~\cite{jax2018github}. 
In contrast, JAX computes derivatives of a function without manually or symbolically differentiating the function, which helps our tool support arbitrary systems and barrier functions, provided that the barrier functions used for control have relative-degree\footnote{A function $p: \mathbb R_+ \times \mathbb R^n \rightarrow \mathbb R$ is said to be of relative-degree $r$ with respect to the dynamics \eqref{sys:ode} if $r$ is the number of times $p$ must be differentiated before one of the control inputs $u$ appears explicitly.} one with respect to the system dynamics. 
If the barrier function of interest has a relative-degree greater than one with respect to the system dynamics, our \texttt{rectify-relative-degree} module may be used to derive a new barrier function whose zero super-level set is a subset of that associated with the original barrier function. The module works by iteratively differentiating the original barrier function with respect to the system dynamics until the control input appears explicitly (as determined by evaluating samples of the term $\frac{\partial h(x_s)}{\partial x}g(x_s)u$ for samples $x_s \in \mathbb \mathcal \mathcal X$), and applying either exponential CBF \cite{nguyen2016exponential} or high-order CBF \cite{xiao2019control} principles to return a ``rectified" barrier function.

The second step in the process is the controller design. One approach to CBF-based control is to pair it with a nominal or reference controller and solve an optimization problem to compute the control solution ``nearest" to the reference input that satisfies the CBF safety constraint(s). For the class of systems we consider, the problem can be posed as a quadratic program (QP) and solved with jaxopt~\cite{blondel2022efficient}, a just-in-time (JIT) compilable Python package for optimization. In our code examples, the QP can be solved in a few milliseconds. 

% \begin{figure}[t]
%   \centering
%   \includegraphics[width=7cm]{fig/object_interaction.pdf}
%   \caption{\texttt{System} and \texttt{Controller} object interaction.}
%   \label{fig:object_interaction}
% \end{figure}

\subsection{Code Generation}

To expedite the construction of new simulations and experiments, we provide a \texttt{codegen} module for creating arbitrary \texttt{dynamics}, \texttt{controller}, and \texttt{cbf} or \texttt{clf} Python functions. These templates provide the user with a direct interface to our \texttt{simulation} module, which supports both single example and multi-processor Monte Carlo trials. An example using \texttt{codegen} to build a simulation is highlighted in Section \ref{subsec.codegen_example}.

\subsection{Interaction with the Robot Operating System (ROS)}

To develop robotics applications, we often use the Robot Operating System (ROS). ROS provides a set of libraries and tools for robotics applications. It provides a message passing infrastructure which enables a publish/subscribe messaging system to allow communication between various components or nodes of the system. For example, we may subscribe to an odometry topic using the \texttt{rospy.Subscriber} function, which triggers a callback function whenever a new message is published to the topic. The callback function can then be used to update the state of the robot. Similarly, we may publish a command to a topic such as angular velocity using the \texttt{rospy.Publisher} function to actuate the robot. 

In our toolbox, we provide a \texttt{ros} module to connect \texttt{dynamics}, \texttt{controller}, \texttt{estimator}, etc. functions with ROS. The module essentially contains wrapper functions to link instantiated functions with ROS topics, which allows other nodes to interact with the system through ROS.
% \todo[inline]{Add simple ROS example -- don't need a full ROS simulation though.}
\textbf{An example may be found at \texttt{something}}.

\subsection{Tutorials}

% \todo[inline]{\textbf{MB / BH}: Update tutorials}

Finally, the toolbox includes a set of step-by-step tutorials for rapid prototyping and benchmark creation of CBF control laws for nonlinear dynamical systems. The tutorials are developed as Jupyter notebooks and walk users through the process step-by-step, from formulating the system using \texttt{codegen} to generating a reference trajectory, designing a barrier function, and relating it to the system dynamics via mathematical derivation. 
Once this is done, the task can be set as a Quadratic Program (QP) with the goal of achieving the desired outcome.
Table \ref{tb:tutorials} summarizes the available tutorials distributed with \cbfkit{}.

\begin{table}[ht]
\centering
\caption{Tutorials on CBF Synthesis}
\label{tb:tutorials}
\begin{tabular}{llc}
\toprule
\multicolumn{1}{c}{\textbf{System}} & \multicolumn{1}{c}{\textbf{Environment}} & \multicolumn{1}{c}{\textbf{CBF order}} \\
\midrule
% Single-delay equation & Static & 1 \\
% Single-delay equation & Dynamic & 1 \\
% Nonlinear unicycle & Static & 2 \\
% Nonlinear unicycle & Dynamic & 2 \\
% Omnidirectional robots & Static & 1 \\
Van der Pol oscillator & Static & 1 \\
Nonlinear unicycle & Static & 2 \\
Nonlinear unicycle & Dynamic & 2 \\
\bottomrule
\end{tabular}
\end{table}

\subsection{Example}\label{subsec.codegen_example}
Consider a unicycle seeking to reach a goal set while avoiding a static, circular obstacle. The code in this subsection is taken directly from \texttt{tutorials/unicycle\_reach\_avoid.py}.

The \texttt{codegen} module is used to generate code for the unicycle equations of motion, the nominal control law, and the barrier functions as shown in Listing \ref{code:codegen}.

\begin{lstlisting}[
    language=Python,
    caption=Codegen for dynamics{,} nominal control{,} and barrier functions.,
    label=code:codegen,
    linerange={3-29}
]
import jax.numpy as jnp
from cbfkit.codegen.generate_model import generate_model


def compute_theta_d(x, y, th):
    thd = f"arctan2(yg - {y}, xg - {x})"
    return f"{th} + arctan2(sin({thd} - {th}), cos({thd} - {th}))"


params = {}
x, y, v, th = "x[0]", "x[1]", "x[2]", "x[3]"
drift = [f"{v} * cos({th})", f"{v} * sin({th})", "0", "0"]
control_mat = ["[0, 0]", "[0, 0]", "[1, 0]", "[0, 1]"]
barriers = [f"({x} - xo)**2 + ({y} - yo)**2 - r**2", f"l**2 - {v}**2"]
params["cbf"] = [{"xo: float": 1.0, "yo: float": 1.0, "r: float": 1.0}, {"l: float": 1.0}]
u_nom = f"kp * (norm([{x} - xg, {y} - yg]) - {v}), kp * ({compute_theta_d(x, y, th)} - {th})"
params["controller"] = {"kp: float": 1.0, "xg: float": 1.0, "yg: float": 1.0}

generate_model(
    directory="./tutorials/models",
    model_name="accel_unicycle",
    drift_dynamics=drift,
    control_matrix=control_mat,
    barrier_funcs=barriers,
    nominal_controller=u_nom,
    params=params,
)
\end{lstlisting}

The \texttt{codegen} module uses formatted strings extensively, which is why all the above formulae are defined as strings (or lists of strings). 
Since the state vector appears in functions as the variable \texttt{x} throughout the toolbox, it is important that the states are defined as ``x[0]", ``x[1]",...,``x[n-1]" (line 11). 
The \texttt{generate\_model} function generates a new folder at \texttt{tutorials/models} called \texttt{accel\_unicycle}, and populates it with files for the plant model (\texttt{plant.py}), the nominal control law (\texttt{controller\_1.py} in \texttt{accel\_unicycle/controllers}), and the barrier functions (\texttt{barrier\_1.py}, \texttt{barrier\_2.py} in \texttt{accel\_unicycle/certificate\_functions} \texttt{/barrier\_functions}).

These modules are imported in Listing \ref{code:instantiation} and used to define \texttt{dynamics} and \texttt{nominal\_controller} functions for the simulation.

\begin{lstlisting}[
    language=Python,
    caption=Instantiating dynamics and nominal controller functions.,
    label=code:instantiation,
    linerange={34-41}
]
import models.accel_unicycle as unicycle
from models.accel_unicycle.certificate_functions.barrier_functions.barrier_1 import cbf
from models.accel_unicycle.certificate_functions.barrier_functions.barrier_2 import cbf2_package

initial_state = jnp.array([2.0, 2.0, 0.0, -3 * jnp.pi / 4])
actuation_limits = jnp.array([1.0, jnp.pi])
dynamics = unicycle.plant()
nominal_controller = unicycle.controllers.controller_1(kp=1.0, xg=-2.0, yg=-2.0)
\end{lstlisting}

In Listing \ref{code:cbf-controller}, the CBF-based controller is built. The important components of this block are i) using the \texttt{rectify\_relative\_degree} function to extract a constraint function with relative-degree 1 w.r.t. the unicycle dynamics from the original obstacle avoidance constraint (line 8-10), ii) concatenating the cbf packages into a usable object using the \texttt{concatenate\_certificates} function (lines 12-15), and iii) instantiating the CBF-based controller function \texttt{controller} (lines 16-22).

\begin{lstlisting}[
    language=Python,
    caption=Instantiating CBF-based controller function.,
    label=code:cbf-controller,
    linerange={45-66}
]
from cbfkit.controllers.model_based.cbf_clf_controllers import vanilla_cbf_clf_qp_controller
from cbfkit.controllers.utils.barrier_conditions.zeroing_barriers import linear_class_k
from cbfkit.controllers.utils.certificate_packager import concatenate_certificates
from cbfkit.controllers.model_based.cbf_clf_controllers.utils.rectify_relative_degree import (
    rectify_relative_degree,
)

cbf1_package = rectify_relative_degree(
    cbf(xo=0.9, yo=1.0, r=0.5), dynamics, len(initial_state), roots=-1.0 * jnp.ones((2,))
)

barriers = concatenate_certificates(
    cbf1_package(certificate_conditions=linear_class_k(10.0)),
    cbf2_package(certificate_conditions=linear_class_k(1.0), l=1.0),
)
controller = vanilla_cbf_clf_qp_controller(
    actuation_limits,
    nominal_controller,
    dynamics,
    barriers,
    p_mat=jnp.diag(jnp.array([1.0, 0.1])),
)
\end{lstlisting}

Finally, the simulation is executed in Listing \ref{code:simulation} from \texttt{initial\_state} and lasts 10 sec at a time-step of 0.01 sec. Additional imports are required, including the \texttt{simulator} module itself and a \texttt{sensor}, \texttt{estimator}, and \texttt{integrator} (lines 1-4). When the simulation is executed by calling \texttt{simulator.execute}, the results for the state, control, state estimate, and state covariance matrix are stored in variables \texttt{x, u, z} and \texttt{p} as \texttt{numpy.ndarray} objects and are immediately available for plotting/analysis.

\begin{lstlisting}[
    language=Python,
    caption=Executing the unicycle simulation.,
    label=code:simulation,
    linerange={71-85}
]
from cbfkit.simulation import simulator
from cbfkit.sensors import perfect
from cbfkit.estimators import naive
from cbfkit.integration import forward_euler

x, u, z, p, dkeys, dvals = simulator.execute(
    x0=initial_state,
    dt=1e-2,
    num_steps=1000,
    dynamics=dynamics,
    integrator=forward_euler,
    controller=controller,
    sensor=perfect,
    estimator=naive,
)
\end{lstlisting}

\section{Experiments}

\subsection{Human Support Robot (HSR)}
\begin{figure}[htbp]
\centerline{\includegraphics[width=5cm]{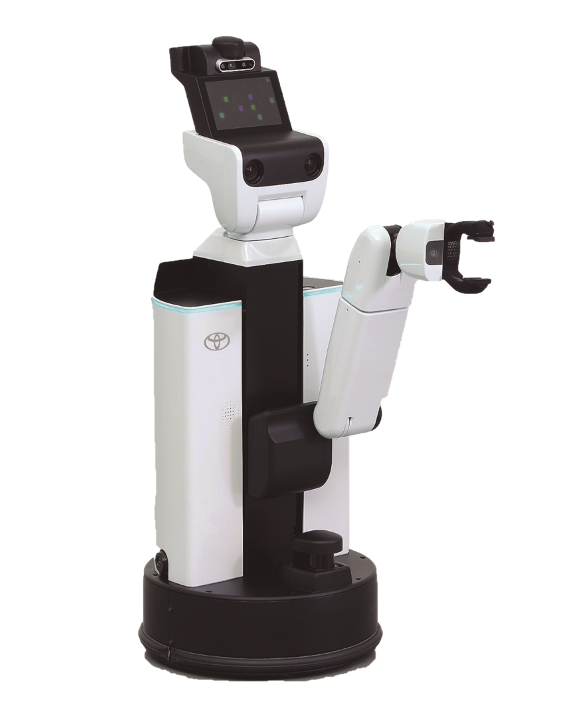}}
\caption{Human Support Robot (HSR). Cr: TOYOTA}
\label{fig:HSR}
\end{figure}

The HSR is a robot
developed by Toyota Motor Corporation to operate in diverse
environments while interacting with humans (Figure \ref{fig:HSR}). It has
a differential wheeled base and five degrees-of-freedom arm.
With its highly maneuverable, compact, and lightweight cylindrical body and folding arm, the HSR can pick objects up off the floor, retrieve objects from shelves, and perform a variety of other tasks. 

\begin{table}[ht]
    \centering
    \caption{Dimensions and Specifications of the Human Support Robot}
    \begin{tabular}{lcc}
    \toprule
    \textbf{Specification} & \textbf{Value} \\
    \midrule
    Body diameter & 430 mm \\
    Body height & 1,005 mm - 1,350 mm \\
    Weight & Approx. 37 kg\\
    Arm length & Approx. 600 mm\\
    Shoulder height & 340 mm - 1,030 mm\\
    Objects that can be held & 1.2 kg or less, 130 mm wide or less\\
    Maximum speed & 0.8 km/h\\
    \bottomrule
    \end{tabular}
    \end{table}

\begin{figure*}[t]
\centering
\resizebox{14cm}{!}{\includegraphics[trim={2.5cm 0mm 3cm 0mm},clip]{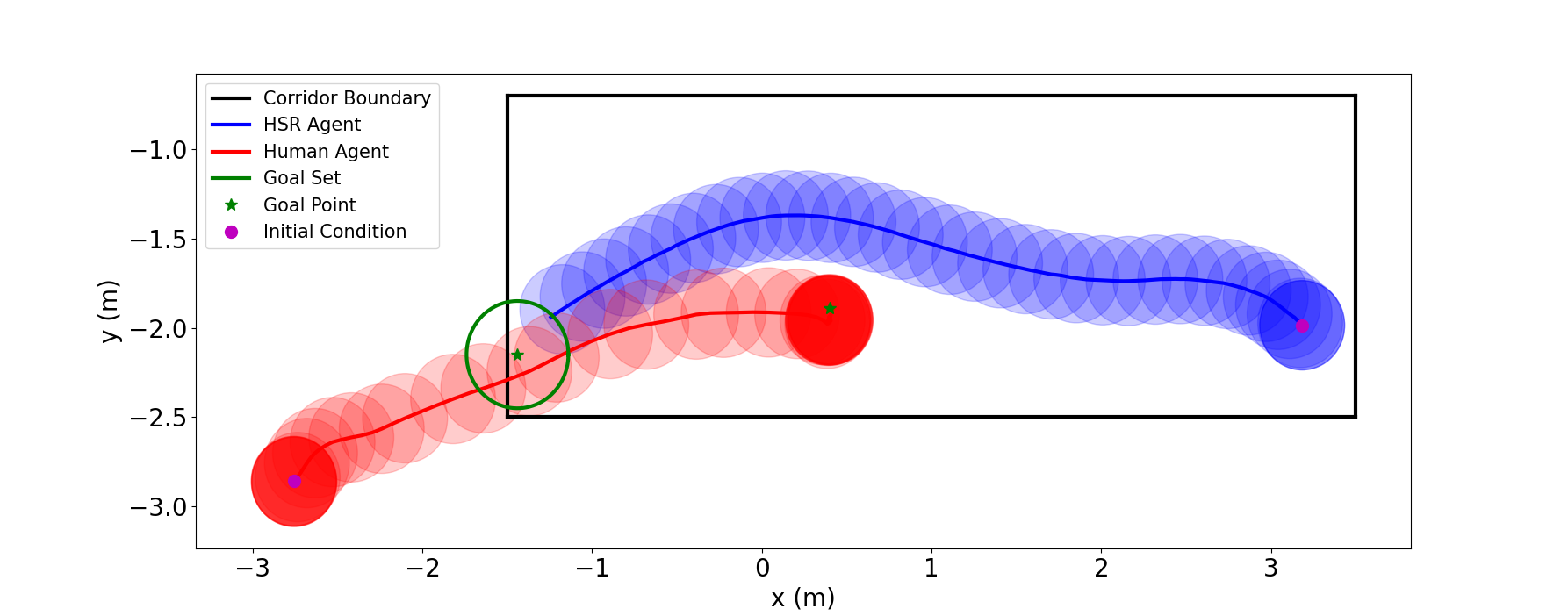}}
\caption{$xy$ paths of the HSR and human agents taken in the goal-reaching experiment. The overlaid circles represent the specified sizes of the ego and human agents and represent temporal evolution at $0.25$ sec increments. An animation of the system behavior can be found at: \url{https://youtu.be/MXQAK2jwLLE}.}
\label{fig:experiment_paths}
\end{figure*}

\begin{figure}[t]
  \centering
  \includegraphics[width=8cm]{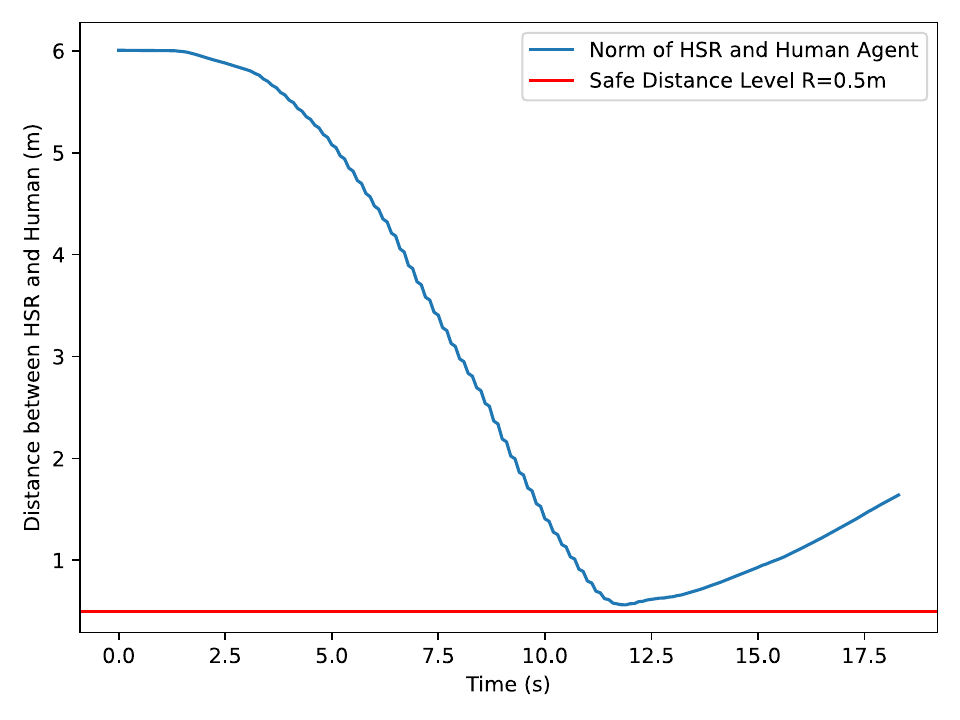}
  \caption{Distance between HSR and Human Agents.}
  \label{fig:dist_hsr_agent}
\end{figure}    

To demonstrate the use of the \cbfkit{} to control a physical system, we conducted an experiment in a laboratory environment in which an HSR was tasked with reaching a goal location while avoiding collisions with a careless human agent and remaining within a specified corridor. In software, the HSR was taken to be the ego agent and assumed to evolve according to the following dynamic unicycle model:
\begin{subequations}\label{eq:unicycle_model}
\begin{align}
    \dot x_e &= v_e\cos\theta _e, \\
    \dot y_e &= v_e\sin\theta _e, \\
    \dot v_e &= a_e, \\
    \dot \theta _e &= \omega _e,
\end{align}
\end{subequations}
where $x_e$ and $y_e$ are the lateral and longitudinal coordinates (in m) with respect to the origin $\bb{s}_0$ of an inertial frame $\mathcal{F}$, $v_e$ is the speed (in m/s), $\theta_e$ is the heading angle (in rad), $a_e$ is the acceleration (in m/s$^2$) in the heading direction, and $\omega_e$ is the heading angular velocity (in rad/s), such that the state vector is $\bb{x}_e = [x_e \; y_e \; v_e \; \theta_e]^\top$ and the control input vector is $\bb{u}_e = [a_e \; \omega_e]^\top$. The human agent was modeled as a 2D single-integrator, i.e.,
\begin{subequations}\label{eq:single_integrator}
\begin{align}
    \dot x_h &= v_{x,h}, \\
    \dot y_h &= v_{y,h},
\end{align}
\end{subequations}
where the state is $\bb{x}_h = [x_h \; y_h]^\top$ and the control input is $\bb{u}_h = [v_{x,h} \; v_{y,h}]^\top$, where $v_{x,h}$ and $v_{y,h}$ are the velocities of the human with respect to the $x$- and $y$-directions.
To ensure collision avoidance, we used a form of the future-focused CBFs introduced for autonomous vehicle control in \cite{BlackJSP2023acc}, in this case defined as
\begin{equation*}
    h_{ff}(\bb{x}_e,\bb{x}_h) = \min_{\tau \in \mathcal{T}}\hat D^2(\bb{x}_e,\bb{x}_h,\tau) - (2R)^2,
\end{equation*}
where $\mathcal{T} = [t,t+T]$ for some time headway $0<T<\infty$, $\hat D: \R^4 \times \R^4 \times \mathcal{T} \rightarrow \R_+$ is the predicted distance between agents under zero-control policies, i.e., $a_i(\tau) = \omega_i(\tau) = 0$, $\forall \tau \in \mathcal{T}$, $i \in \{e,h\}$, and $R>0$ is a safe distance. To enforce that the HSR remained inside the specified corridor, we defined an admissible rectangular region with edges $x_{min},x_{max},y_{min},y_{max} \in \R$ and used the following CBFs:
\begin{align*}
    h_{r1}(\bb{x}_e) &= v_e\cos\theta_e + x_e - x_{min}, \\
    h_{r2}(\bb{x}_e) &= x_{max} - x_e - v_e\cos\theta_e, \\
    h_{r3}(\bb{x}_e) &= v_e\sin\theta_e + y_e - y_{min}, \\
    h_{r4}(\bb{x}_e) &= y_{max} - y_e - v_e\sin\theta_e.
\end{align*}
We then used a form of the well-known CBF quadratic program (CBF-QP) control law to control the HSR, described as follows:
\begin{subequations}\label{eq:cbf_qp_controller}
\begin{align}
    \bb{u}_e^* = \argmin_{\bb{u}_e \in \mathcal{U}_e} \frac{1}{2}\|\bb{u}_e&-\bb{u}_e^0\|^2 \label{subeq.cost_fcn}\\
    \nonumber \textrm{s.t.}& \quad \forall k \in \{ff, r1, r2,r3, r4\} \\
    L_fh_k + L_gh_k\bb{u}_e &\geq -\alpha(h_k), \label{subeq.cbf_constraint}
\end{align}
\end{subequations}
where \eqref{subeq.cost_fcn} seeks to minimize the deviation of the control solution $\bb{u}_e^*$ from the reference input $\bb{u}_e^0$, and \eqref{subeq.cbf_constraint} enforces the CBF-based safety condition for each of the specified constraint functions. In this experiment, we used a reference control $\bb{u}_e^0$ based on the LQR control law found in \cite[App. I]{BlackJSP2023acc}, and chose $\alpha(y) = y$.

The measured paths for the HSR and human agents are shown in Figure \ref{fig:experiment_paths}. Evidently, the CBF-QP controller \eqref{eq:cbf_qp_controller} steers the HSR out of the path of the human agent in advance of any dangerous scenario despite the HSR's nominal LQR controller seeking to take it directly through the human agent en route to the goal. As shown in Figure \ref{fig:dist_hsr_agent}, the HSR obeys the required safety margin with respect to the human agent at all times.

% \begin{figure}[!t]
%     \centering
%         \includegraphics[width=1\columnwidth,trim={20mm 2mm 20mm 5mm},clip]{hsr_experiment.png}
%     \caption{\small{$xy$ paths of the HSR and human agents taken in the goal-reaching experiment}.}}\label{fig:experiment_paths}
% \end{figure}

\section{Conclusion}

In this paper, we presented the first publicly available open source Python toolbox for CBF-based control for generic systems both in simulation and ROS-enabled hardware experimentation.
The toolbox, called \cbfkit{}, is developed using a functional programming architecture so it promotes immutability and reliability.
The toolbox already contains a number of CBF methods from the literature, and it has been successfully applied for motion planning on the Toyota HSR in a laboratory environment.
In the near future, we plan to extend \cbfkit{} to include a variety of methods and algorithms from the literature.
We also envision the toolbox to provide a number of benchmarks that will enable the research community to compare different algorithms and approaches.

\bibliographystyle{IEEEtran}
\bibliography{main}

\end{document}